\documentclass{article}

\usepackage{microtype}
\usepackage{graphicx}
\usepackage{subfigure}
\usepackage{booktabs} 
\usepackage{amsmath}
\usepackage{amssymb}  
\usepackage{wrapfig}
\usepackage{algorithm}

\usepackage{hyperref}

\usepackage[accepted]{icml2018}

\icmltitlerunning{Meta-Reinforcement Learning of Structured Exploration Strategies}

\begin{document}

\twocolumn[
\icmltitle{Meta-Reinforcement Learning of Structured Exploration Strategies}

\icmlsetsymbol{equal}{*}

\begin{icmlauthorlist}
\icmlauthor{Abhishek Gupta}{berkeley}
\icmlauthor{Russell Mendonca}{berkeley}
\icmlauthor{YuXuan Liu}{berkeley}
\icmlauthor{Pieter Abbeel}{berkeley,embodied}
\icmlauthor{Sergey Levine}{berkeley}
\end{icmlauthorlist}


\icmlaffiliation{berkeley}{UC Berkeley, Berkeley, CA, USA\\}
\icmlaffiliation{embodied}{Embodied Intelligence, USA\\}

\icmlcorrespondingauthor{Abhishek Gupta}{abhigupta@eecs.berkeley.edu}

\icmlkeywords{Exploration, Meta-learning, Adaptation, Robotics}

\vskip 0.3in
]

\printAffiliationsAndNotice{}

\begin{abstract}
Exploration is a fundamental challenge in reinforcement learning (RL). Many of the current exploration methods for deep RL use task-agnostic objectives, such as information gain or bonuses based on state visitation. However, many practical applications of RL involve learning more than a single task, and prior tasks can be used to inform how exploration should be performed in new tasks. In this work, we explore how prior tasks can inform an agent about how to explore effectively in new situations. We introduce a novel gradient-based fast adaptation algorithm -- model agnostic exploration with structured noise (MAESN) -- to learn exploration strategies from prior experience. The prior experience is used both to initialize a policy and to acquire a latent exploration space that can inject structured stochasticity into a policy, producing exploration strategies that are informed by prior knowledge and are more effective than random action-space noise. We show that MAESN is more effective at learning exploration strategies when compared to prior meta-RL methods, RL without learned exploration strategies, and task-agnostic exploration methods. We evaluate our method on a variety of simulated tasks: locomotion with a wheeled robot, locomotion with a quadrupedal walker, and object manipulation.
\end{abstract}

\section{Introduction}
\label{sec:intro}

Deep reinforcement learning methods have been shown to learn complex tasks ranging from games~\citep{Atari} to robotic control~\citep{Levine16,ddpg} with minimal supervision, by simply exploring the environment and experiencing rewards. As the task becomes more complex or temporally extended, more na\"{i}ve exploration strategies become less effective. Devising more effective exploration methods is therefore a critical challenge in reinforcement learning. Prior works have proposed guiding exploration based on criteria such as intrinsic motivation ~\cite{schmidhuber, bradly_acvp, singh}, state-visitation counts ~\cite{lopes, MBIE-EB, pseudocounts}, Thompson sampling and bootstrapped models ~\cite{thompsonsampling, bootstrappedDQN}, optimism in the face of
uncertainty~\cite{R-max, E3}, and parameter space exploration ~\cite{parameterspacenoise, noisynets}.

These exploration strategies are largely task agnostic, in that they aim to provide good exploration without exploiting the particular structure of the task itself. However, an intelligent agent interacting with the real world will likely need to learn many tasks, not just one, in which case prior tasks can be used to inform how exploration in new tasks should be performed. For example, a robot that is tasked with learning a new household chore likely has prior experience of learning other related chores. It can draw on these experiences in order to decide how to explore the environment to acquire the new skill more quickly. Similarly, a walking robot that has previously learned to navigate different buildings doesn't need to reacquire the skill of walking when it must learn to navigate through a maze, but simply needs to explore in the space of navigation strategies. 

In this work, we study how experience from multiple distinct but related prior tasks can be used to autonomously acquire directed exploration strategies via meta-learning. Meta-learning, or learning to learn, refers to the problem of learning strategies for fast adaptation by using prior tasks~\cite{schmidhuber,thrun,hochreiter,matching}. Several methods have aimed to address meta-learning in RL contexts~\cite{rl2,learning_to_rl} by training recurrent models that ingest past states, actions, and rewards, and predict new actions that will maximize rewards for the task at hand. These methods are not ideal for learning to explore, as we illustrate both conceptually and empirically. There are two main reasons for this. First, good exploration strategies are qualitatively different from optimal policies: while an optimal policy is typically deterministic in fully observed environments, exploration depends critically on stochasticity. Methods that simply recast the meta-RL problem into an RL problem~\cite{rl2,learning_to_rl} generally acquire behaviors that exhibit insufficient variability to explore effectively in new settings for difficult tasks. The same policy has to represent highly exploratory behavior \emph{and} adapt very quickly to optimal behavior, which becomes very difficult with typical time-invariant representations for action distributions. Second, many current meta-RL methods aim to learn the entire ``learning algorithm,'' for example by using a recurrent model. While this allows them to adapt very quickly, via a single forward pass of the RNN, it greatly limits their asymptotic performance when compared to learning from scratch, since the learned ``algorithm'' (i.e., RNN) generally does not correspond to a convergent iterative optimization procedure, unlike a standard RL method.

We aim to address both of these challenges by devising a meta-RL algorithm that adapts to new tasks by following the policy gradient, while also injecting learned structured stochasticity into a latent space to enable effective exploration. Our algorithm, which we call model agnostic exploration with structured noise (MAESN), uses prior experience both to initialize a policy and to learn a latent exploration space from which it can sample temporally coherent structured behaviors, producing exploration strategies that are stochastic, informed by prior knowledge, and more effective than random noise. Importantly, the policy and latent space are \emph{explicitly} trained to adapt quickly to new tasks with the policy gradient. Since adaptation is performed by following the policy gradient, our method achieves at least the same asymptotic performance as learning from scratch (and often performs substantially better), while the structured stochasticity allows for randomized but task-aware exploration.

Our experimental evaluation shows that existing meta-RL methods, including MAML~\citep{maml} and RNN-based algorithms~\cite{rl2,learning_to_rl}, are limited in their ability to acquire complex exploratory policies, likely due to limitations on their ability to acquire a strategy that is both stochastic and structured with policy parameterizations that can only introduce time-invariant stochasticity into the action space. While in principle certain RNN based architectures could capture time-correlated stochasticity, we find experimentally that current methods fall short. Effective exploration strategies must select randomly from among the \emph{potentially useful} behaviors, while avoiding behaviors that are highly unlikely to succeed.
MAESN leverages this insight to acquire significantly better exploration strategies by incorporating learned time-correlated noise through its meta-learned latent space, and training both the policy parameters and the latent exploration space explicitly for fast adaptation. We show that combining these mechanisms together produces a meta-learning algorithm that can learn to explore substantially better than prior meta-learning methods and adapt quickly to new tasks. In our experiments, we find that we are able to explore coherently and adapt quickly for a number of simulated manipulation and locomotion tasks with challenging exploration components.

\section{Related Work}

Exploration is a fundamental problem in RL. While simple methods such as $\epsilon$-greedy or Gaussian exploration are used widely, they are often insufficient for environments with significant complexity and delayed rewards. A wide range of more sophisticated strategies have been proposed based on optimism in the face of uncertainty ~\cite{R-max, E3}, intrinsic motivation~\cite{schmidhuber, bradly, singh}, Thompson sampling~\cite{thompsonsampling, bootstrappedDQN}, information gain~\cite{VIME}, and parameter space exploration ~\cite{parameterspacenoise, noisynets}. While many of these methods are effective at improving exploration, they are all task agnostic, and therefore do not utilize prior knowledge about the world that can be gained from other tasks. MAESN instead aims to incorporate experience from distinct but structurally similar prior tasks to learn exploration strategies. In a spirit similar to parameter-space exploration, MAESN injects temporally correlated noise to randomize exploration strategies, but the way this noise is utilized and sampled is determined by a meta-learning process and informed by past experience.

Our algorithm is based on the framework of meta-learning~\cite{thrun, schmidhuber, bengio92}, which aims to learn models that can adapt quickly to new tasks. Meta-learning algorithms learn optimizers~\cite{ltlbgdbgd}, update rules~\cite{ravi}, or entire RL algorithms~\cite{rl2, learning_to_rl, metacritic}. Such methods have been effective at solving new supervised learning problems with very few samples~\cite{matching, mann, snail, ravi, hochreiter, Koch, snell17, metanetworks}. Our approach is most closely related to model-agnostic meta-learning (MAML)~\cite{maml}, which directly trains for model parameters that can adapt quickly with standard gradient descent. This method has the benefit of allowing for similar asymptotic performance as learning from scratch, especially when the new task differs from the training task, while still enabling acceleration from meta-training. However, our experiments show that MAML alone, as well as prior meta-RL methods, are not as effective at learning to explore, due to their lack of structured stochasticity.

Our proposed method (MAESN) introduces structured stochasticity into meta-learning via a learned latent space. Latent space models for RL have been explored in several prior works, though not in the context of meta-learning or learning exploration strategies~\cite{hausman, carlos, zico}. These methods do not explicitly train for fast adaptation, and comparisons in Section~\ref{sec:experiments} illustrate the advantages of our method.
Concurrently to our work, \citet{bradly} also explores meta-learning for exploration in an unpublished manuscript, but does not introduce structured stochasticity, and presents results that do not show a significant improvement over MAML and other meta-learning algorithms. In contrast, our results show that MAESN substantially outperforms prior meta-learning methods.

\section{Preliminaries}

In this section, we introduce our meta-learning problem formulation, and describe model-agnostic meta-learning, a prior meta-learning algorithm that MAESN builds on.

\subsection{Meta-Learning for Reinforcement Learning}

In meta-RL, we consider a distribution $p(\tau)$ over tasks, where each task $\tau_i$ is a different Markov decision process (MDP) $M_i = (S, A, T_i, R_i)$, with state space $S$, action space $A$, transition distribution $T_i$, and reward function $R_i$. The reward function and transitions vary across tasks. Meta-RL aims to to learn a policy that can adapt to maximize the expected reward for novel tasks from $p(\tau)$ as efficiently as possible. This can be done in a number of ways, using gradient descent based methods~\cite{maml} or recurrent models that ingest past experience~\cite{rl2, learning_to_rl, snail}.

We build on the gradient-based meta-learning framework of MAML~\cite{maml}, which trains a model in such a way that it can adapt quickly with standard gradient descent, which in RL corresponds to the policy gradient. The meta-training objective for MAML can be written as
\begin{align}
\max_{\theta} & \sum_{\tau_i} \mathbb{E}_{\pi_{\theta'}} [\sum_t R_i(s_t)]\\
&\theta' = \theta +  \alpha\mathbb{E}_{\pi_{\theta}} [\sum_t R_i(s_t)\nabla_{\theta}\log\pi_{\theta}(a_t|s_t)]
\end{align}
The intuition behind this optimization objective is that, since the policy will be adapted at meta-test time using the policy gradient, we can optimize the policy parameters so that one step of policy gradient improves its performance on any meta-training task as much as possible.

Since MAML reverts to conventional policy gradient when faced with out-of-distribution tasks, it provides a natural starting point for us to consider the design of a meta-exploration algorithm: by starting with a method that is essentially on par with task-agnostic RL methods that learn from scratch in the \emph{worst} case, we can improve on it to incorporate the ability to acquire stochastic exploration strategies from experience, as discussed in the following section, while preserving at least the same asymptotic performance~\cite{chelseauniversality}.

\section{Model Agnostic Exploration with Structured Noise}

While meta-learning has been shown to be effective for fast adaptation on several RL problems ~\cite{maml,rl2}, the prior methods generally focus on tasks where either a few trials are sufficient to identify the goals of the task~\cite{maml}, or the policy should acquire a consistent ``search'' strategy, for example to find the exit in new mazes~\cite{rl2}. Both of these adaptation regimes differ substantially from stochastic exploration. Tasks where discovering the goal requires exploration that is both stochastic \emph{and} structured cannot be easily captured by such methods, as demonstrated in our experiments. Specifically, there are two major shortcomings with these methods: (1) The stochasticity of the policy is limited to time-invariant noise from action distributions, which fundamentally limits the exploratory behavior it can represent. (2) For RNN based methods, the policy is limited in its ability to adapt to new environments, since adaptation is performed with a forward pass of the recurrent network. If this single forward pass does not produce good behavior, there is no further mechanism for improvement. Methods that adapt by gradient descent, such as MAML, simply revert to standard policy gradient and can make slow but steady improvement in the worst case, but do not address (1). In this section, we introduce a novel method for learning structured exploration behavior based on gradient based meta-learning which is able to learn good exploratory behavior \emph{and} adapt quickly to new tasks that require significant exploration.

\subsection{Overview}

Our algorithm, which we call model agnostic exploration with structured noise (MAESN), combines structured stochasticity with MAML. MAESN is a gradient-based meta-learning algorithm that introduces stochasticity not just by perturbing the actions, but also through a learned latent space. Both the policy and the latent space are trained with meta-learning to provide for fast adaptation to new tasks. When solving new tasks at meta-test time, a different sample is generated from this latent space for each trial, providing structured and temporally correlated stochasticity. The distribution over the latent variables is then adapted to the task via policy gradient updates. We first show how structured stochasticity can be introduced through latent spaces, and then describe how both the policy and the latent space can be meta-trained to form our overall algorithm.

\subsection{Policies with Latent State}

Typical stochastic policies parameterize action distributions using an action distribution $\pi_{\theta}(a|s)$ that is independent for each time step. However, this representation has no notion of temporally coherent randomness throughout the trajectory, since noise is added independently at each time step. This greatly limits its exploratory power, since the policy essentially ``changes its mind'' about what it wants to explore every time step. The distribution $\pi_{\theta}(a|s)$ is also typically represented with simple parametric distributions, such as unimodal Gaussians, which restricts its ability to model task-dependent covariances.
\begin{wrapfigure}{r}{0.4\columnwidth}
  \centering
  \includegraphics[height=\linewidth]{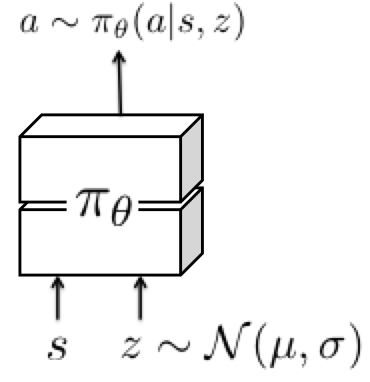}
  \caption{Neural network parameterization of a policy conditioned on latent variable $z \sim \mathcal{N}(\mu, \sigma)$ which is sampled once per episode. Actions are sampled once per time-step from $\pi_{\theta}(a|s,z)$}
\end{wrapfigure}
To incorporate temporally coherent exploration behavior and allow the policy to model more complex time-correlated stochastic processes, we can condition the policy on per-episode random variables drawn from a learned latent distribution. Since the latent variables are sampled only once per episode, they provide temporally coherent stochasticity. Intuitively, the policy decides only once what it will try for that episode, and then sticks to this plan. Furthermore, since the random sample is provided as an input, a nonlinear neural network policy can transform this random variable into arbitrarily complex distributions. The resulting policies can be written as $\pi_{\theta}(a|s, z)$, where $z \sim \mathcal{N}(\mu, \sigma)$ and $\mu$ and $\sigma$ are learnable parameters. This structured stochasticity can provide more coherent exploration, by sampling entire behaviors or goals, rather than simply relying on independent random actions. Related policy representations have been explored in prior work ~\cite{hausman, carlos}. However, we take this design a step further by meta-learning the latent space for efficient adaptation.

\begin{figure}[!h]
\centering
\includegraphics[height=0.5\linewidth]{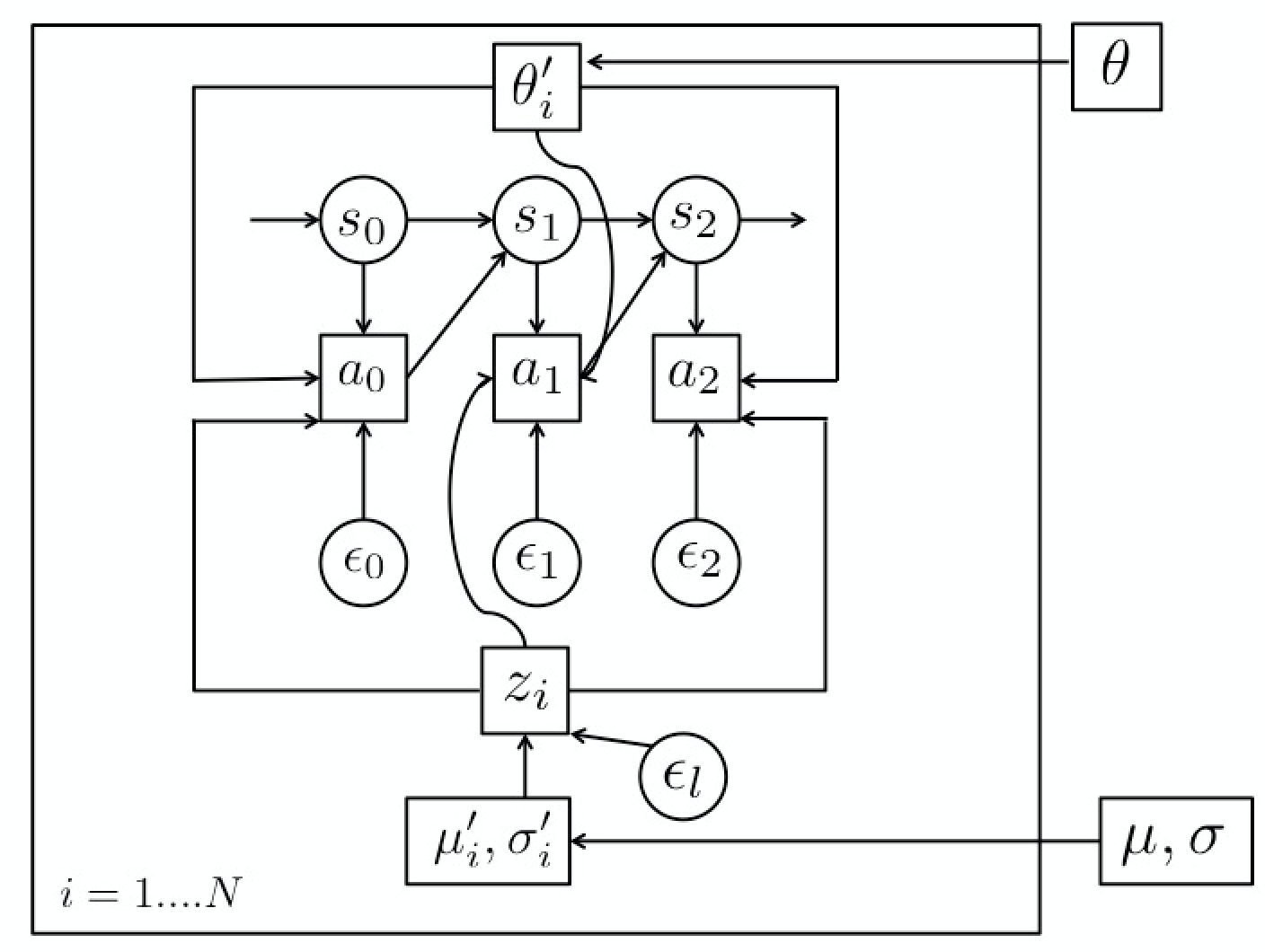}
\caption{Computation graph for MAESN. There are N tasks, each with a set of latent distribution parameters $\mu_i', \sigma_i'$ and policy parameters $\theta'$, updated in the inner loop of meta-learning from the overall parameters $\theta$ and pre-update variational parameters $\mu_i, \sigma_i$. The sampling procedure for the actions introduces time correlated noise by conditioning the policy on a latent variable $z_i$ which is kept fixed through the episode. The action is still drawn per time-step but overall exploration is time-correlated due to $z_i$.
}
\label{fig:computation_graph}
\end{figure}

\subsection{Meta-Learning Latent Variable Policies}

Given a latent variable conditioned policy as described above, our goal is to train it so as to capture coherent exploration strategies that enable fast adaptation on new tasks. We use a combination of variational inference and gradient based meta-learning to achieve this. Specifically, our aim is to meta-train the policy parameters $\theta$ so that they can make use of the latent variables to perform coherent exploration on a new task and adapt as fast as possible. To that end, we learn a set of latent space distribution parameters for each task for optimal performance \emph{after} a policy gradient adaptation step. This procedure encourages the policy to actually make use of the latent variables for exploration. From one perspective MAESN can be understood as augmenting MAML with a latent space to inject structured noise, from a different perspective it amounts to learning a structured latent space as in ~\cite{hausman} but trained for quick adaptation to new tasks via policy gradient. 

To formalize the objective for meta-training, we introduce per-task variational parameters $\mu_i, \sigma_i$ that define the per-task latent variable distribution $\mathcal{N}(\mu_i, \sigma_i)$, one for each task $\tau_i$. Meta-training involves optimizing the initial policy parameters $\theta$, which are shared for all tasks, and the per-task initial variational parameters $\{(\mu_0, \sigma_0), (\mu_1, \sigma_1)...\}$, such that we maximize expected reward after one policy gradient update. As is standard in variational inference, we also add to the objective the KL-divergence between the Gaussian distribution corresponding to each set of pre-update variational parameters and the latent variable prior, which is simply a unit Gaussian. Intuitively, this means that for every iteration of meta-training, we sample from the latent variable conditioned policies represented by $\theta$, $(\mu_i, \sigma_i)$, perform an ``inner'' gradient update on the variational parameters for each task (and, optionally, the policy parameters) to get the post-update parameters $\theta_i'$ and $(\mu_i', \sigma_i')$, and then meta-update the parameters $\theta$, $(\mu_0, \sigma_0), (\mu_1, \sigma_1)...$ such that the sum of expected task rewards over all tasks using the updated latent-conditioned policies $\theta_i'$, $(\mu_i', \sigma_i')$ is maximized. 

As in MAML~\cite{maml}, this involves differentiating through the policy gradient. During meta-training, the ``inner'' update corresponds to the standard REINFORCE policy gradient, which is straightforward to differentiate~\cite{maml}, while the meta-optimizer is the more powerful Trust Region Policy Optimization(TRPO) algorithm~\cite{TRPO}. A concise description of the meta-training procedure is provided in Algorithm~\ref{algo:maesn}, and the computation graph representing MAESN is shown in Fig~\ref{fig:computation_graph}. The full meta-training problem can be stated as
\begin{align}
    \max_{\theta, \mu_i, \sigma_i} &\sum_{i \in tasks} E_{\substack{a_t \sim \pi(a_t | s_t ; \theta_i', z_i') \\ z_i' \sim \mathcal{N}(\mu_i', \sigma_i')}}[\sum_t R_i(s_t)] - \\ &\sum_{i \in tasks} D_{KL}(\mathcal{N}(\mu_i, \sigma_i) \| \mathcal{N}(0, I)) \\ &\mu_i' = \mu_i +  \alpha_{\mu} \circ \nabla_{\mu_i} E_{\substack{a_t \sim \pi(a_t | s_t ; \theta, z_i) \\ z_i \sim \mathcal{N}(\mu_i, \sigma_i)}}[\sum_t R_i(s_t)] \\
    & \sigma_i' = \sigma_i +  \alpha_{\sigma} \circ \nabla_{\sigma_i} E_{\substack{a_t \sim \pi(a_t | s_t ; \theta, z_i) \\ z_i \sim \mathcal{N}(\mu_i, \sigma_i)}}[\sum_t R_i(s_t)] \\ &\theta_i' = \theta +  \alpha_{\theta} \circ \nabla_{\theta} E_{\substack{a_t \sim \pi(a_t | s_t ; \theta, z_i) \\ z_i \sim \mathcal{N}(\mu_i, \sigma_i)}}[\sum_t R_i(s_t)].
\end{align}
The two objective terms are the post-update expected reward for each task and the KL-divergence between each task's variational parameters and the prior. The $\alpha$ values are per-parameter step sizes, and $\circ$ is an elementwise product. The last update (to $\theta$) is optional. We found that we could in fact obtain better results simply by omitting this update, which corresponds to meta-training the initial policy parameters $\theta$ simply to use the latent space efficiently, without training the parameters themselves explicitly for fast adaptation. Including the $\theta$ update makes the resulting optimization problem more challenging, and we found it necessary to employ a stage-wise training procedure in this case, where the policy is first meta-trained without the $\theta$ update, and then the $\theta$ update is added. However, even in this case, the policy's fast adaptation performance does not actually improve over simply omitting this update during meta-training, so we do not use this procedure in our experiments. We also found that meta-learning the step size $\alpha$ separately for each parameter was crucial to achieve good performance.

Note that exploration in the inner loop happens both via exploration in the action space as well as in the latent space, but the latent space exploration is temporally coherent. The MAESN objective enables structured exploration through the noise in latent space, while explicitly training for fast adaptation via policy gradient. We could in principle train such a model without meta-training for adaptation, which resembles the model proposed by \citet{hausman}, but we will show in our experimental evaluation that our meta-training produces substantially better results.

Interestingly, during the course of meta-training, we find that the variational parameters $(\mu_i,\sigma_i)$ for each task are usually close to the prior at convergence, in contrast to the non-meta-training approach, where the model is trained to maximize expected reward without a policy gradient update. This has a simple explanation: meta-training optimizes for \emph{post-update} rewards, after $(\mu_i,\sigma_i)$ have been updated, so even if $(\mu_i,\sigma_i)$ initially matches the prior, it does not match the prior after an update. This allows us to succeed on new tasks at meta-test time for which we do not have a good initialization for $(\mu,\sigma)$, and have no choice but to begin with the prior, as discussed in the next section.

This objective is a version of the variational evidence lower bound (ELBO), which is typically written as
\begin{equation}
    \log p(x) \geq E_q[\log p(x|z)] - D_{KL}(q \| p).
\end{equation}
MAESN performs variational inference, where the likelihood $\log p(x|z)$ is the reward  $R$ under the updated policy $\theta_i'$ and latent variational parameters $\mu_i', \sigma_i'$. Although it might seem unconventional to treat reward values as likelihoods, this can be made formal via an equivalence between entropy-maximizing RL and probabilistic graphical models. A full derivation of this equivalence is outside of the scope of this work, and we refer the reader to prior work for details~\cite{softq, lmdp}.

Note that MAESN not only introduces exploration through the latent space, but also optimizes the policy parameters and latent space such that the task reward is maximized after one gradient update. This ensures fast adaptation when learning new tasks with policy gradients.
 
 \begin{algorithm}[b]
\begin{algorithmic}[1]
 \STATE{Initialize variational parameters $\mu_i, \sigma_i$ for each training task $\tau_i$}
\FOR{iteration $k \in \{1,\dots,K\}$}
    \STATE{Sample a batch of $N$ training tasks from $p(\tau)$}
    \FOR{task $\tau_i \in \{1,\dots,N\}$}
        \STATE{Gather data using the latent conditioned policy $\theta$, $(\mu_i, \sigma_i)$}
        \STATE{Compute inner policy gradient on variational parameters via Equation (4) and (5) (optionally (6))}
    \ENDFOR
    \STATE{Compute meta update on both latents and policy parameters by optimizing (3) with TRPO}
\ENDFOR
\end{algorithmic}
\caption{MAESN meta-RL algorithm}
\label{algo:maesn}
\end{algorithm}
 
\subsection{Using the Latent Spaces for Exploration}
\label{sec:adaptation}
Let us consider a new task $\tau_i$ with reward $R_i$, and a learned model with policy parameters $\theta$ and variational parameters $\mu_i, \sigma_i$. For exploration in this task, we can initialize the latent distribution to the prior $\mathcal{N}(\mu, \sigma) = p(z) = \mathcal{N}(0, I)$, since the KL regularization drives the variational parameters to the prior during meta-training. Adaptation to the new task is then done by simply using the policy gradient to adapt $\mu$ and $\sigma$ via backpropagation on the RL objective,
\begin{equation}
\max_{\mu, \sigma} E_{\substack{a_t \sim \pi(a_t | s_t, \theta, z) \\ z \sim \mathcal{N}(\mu, \sigma)}}[\sum_t R(s_t)],
\end{equation}
where $R$ represents the sum of rewards along the trajectory. Since we meta-trained to adapt $\mu, \sigma$ in the inner loop, we adapt these parameters at meta-test time as well. To compute the gradients with respect to $\mu, \sigma$, we need to backpropagate through the sampling operation $z \sim \mathcal{N}(\mu, \sigma)$, using either likelihood ratio or the reparameterization trick. The likelihood ratio update is
\begin{equation}
\vspace*{-10pt}
\nabla_{\mu, \sigma} \eta = E_{\substack{a_t \sim \pi(a_t | s_t ; \theta, z) \\ z \sim \mathcal{N}(\mu, \sigma)}}[\nabla_{\mu, \sigma}\log p_{\mu, \sigma}(z)\sum_t R(s_t)].
\end{equation}

One final detail with meta-learning exploration strategies is the question of rewards. While our goal is to adapt quickly with sparse and delayed rewards at meta-test time, this goal poses a major challenge at meta-training time: if the tasks themselves are too difficult to learn from scratch, they will also be difficult to solve at meta-training time, making it hard for the meta-learner itself to make progress. While this issue could potentially be addressed by using many more samples or existing task-agnostic exploration strategies during meta-training only, an even simpler solution is to introduce some amount of reward shaping during meta-training (both for our method and for baselines). As we will show in our experiments, exploration strategies meta-trained with reward shaping actually generalize effectively to sparse and delayed rewards, despite the mismatch in the reward function. This can be viewed as a kind of mild instrumentation of the meta-training setup, and could be replaced with other mechanisms, such as very large sample sizes and task-agnostic exploration bonuses, in future work.

\section{Experiments}
\label{sec:experiments}

Our experiments aim to comparatively evaluate our meta-learning method and study the following questions: (1) Can meta-learned exploration strategies with structured noise explore coherently and adapt quickly to new tasks, providing a significant advantage over learning from scratch? (2) How does meta-learning with MAESN compare with prior meta-learning methods such as MAML~\cite{maml} and RL2~\cite{rl2}, as well as latent space learning methods~\cite{hausman}? (3) Can we visualize the exploration behavior and see coherent exploration strategies with MAESN? (4) Can we better understand which components of MAESN are the most critical?
\subsection{Task Setup}

We evaluated our method on three task distributions $p(\tau)$. For each family of tasks we used 100 distinct meta-training tasks, each with a different reward function $R_i$. After meta-training on a particular distribution of tasks, MAESN is able to explore well and adapt quickly to tasks drawn from this distribution (with sparse rewards). We describe the three task distributions, and associated exploration challenges.

\begin{figure}
    \centering
    \includegraphics[width=0.25\linewidth]{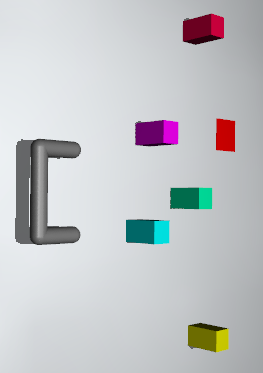}
    \includegraphics[width=0.5\linewidth]{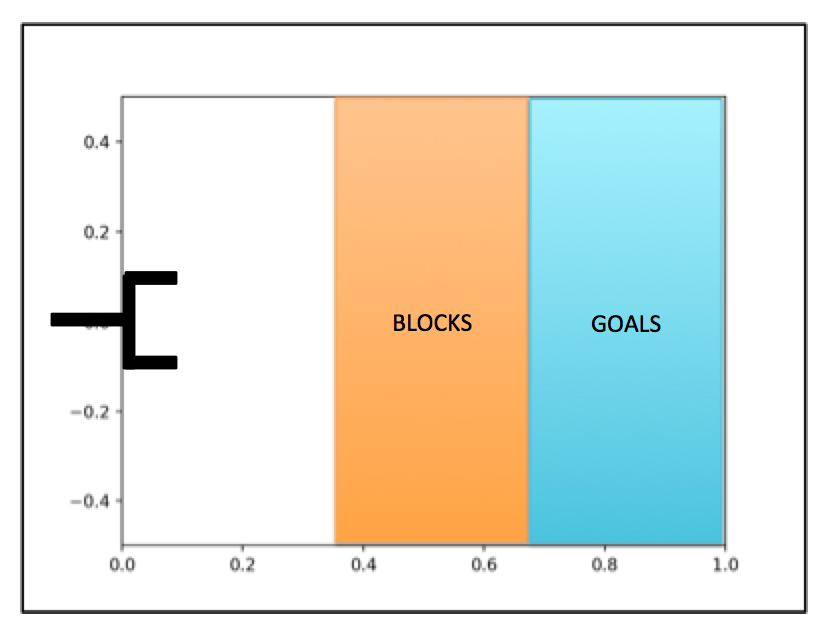}
    \caption{\textbf{Left}: Object manipulation with a robotic gripper pushing various colored blocks to the red goal square. \textbf{Right}: Distribution of blocks (orange region) and goals (blue region) across the task distribution indicating task diversity}
    \label{fig:blockpush}
    \vspace*{-15pt}
\end{figure}

\paragraph{Robotic Manipulation.}
The goal in these tasks is to push blocks to target locations with a robotic hand. In each task, several blocks are placed at random positions. Only one block (unknown to the agent) is relevant for each task, and that block must be moved to a goal location (see Fig.~\ref{fig:blockpush}). The different tasks in the distribution require pushing different blocks from different positions and to different goals. The state consists of the positions of all of the objects, the goal position, and the configuration of the hand. During meta-training, the reward function corresponds to the negative distance between the relevant block and the goal, but during meta-testing, only a sparse reward function for reaching the goal with the correct goal is provided. A coherent exploration strategy should pick random blocks to move to the goal location, trying different blocks on each episode to discover the right one. This task is generally representative of exploration challenges in robotic manipulation: while a robot might perform a variety of different manipulation skills, generally only motions that actually interact with objects in the world are useful for coherent exploration. We use full state representation for this task, leaving vision based policies to future work.

\begin{figure}[!h]
\centering
    \includegraphics[width=0.22\linewidth]{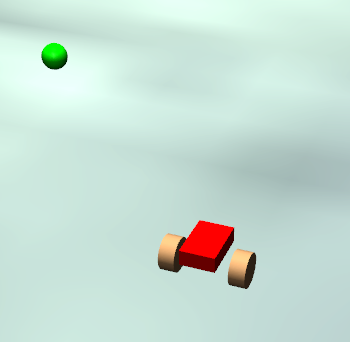}
    \includegraphics[width=0.33\linewidth]{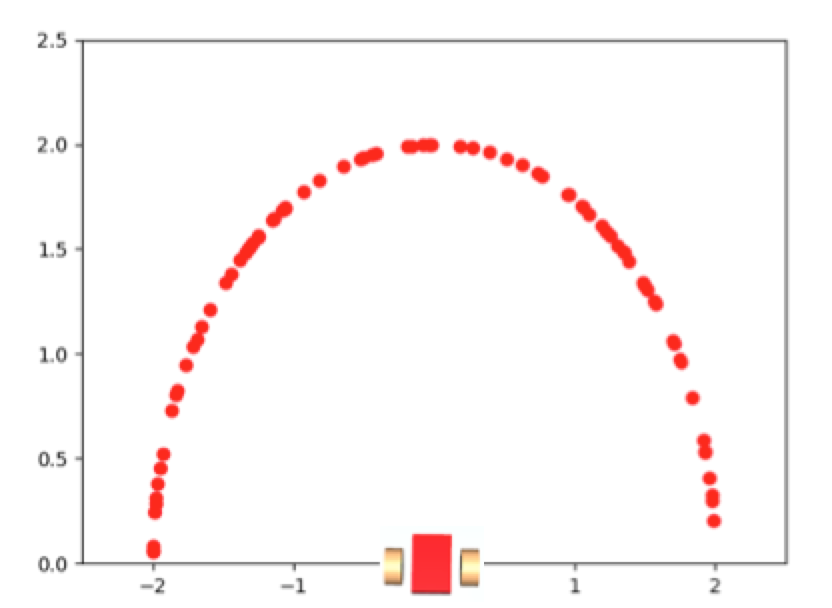}
    \vspace*{-10pt}
    \caption{\textbf{Left:} Locomotion with a wheeled robot navigating to the goal depicted by the green sphere, \textbf{Right:} Depiction of the distribution of tasks. The red points indicate various goals the agent might need to reach, each of which is a task from $p(\tau)$}
\label{fig:wheeled}
\end{figure}

\paragraph{Wheeled Locomotion.}
For our second task distribution we consider a wheeled robot that has to navigate to different goals. Each task has a different goal location. The robot controls its two wheels independently, using them to move and turn. This task family is illustrated in Fig.~\ref{fig:wheeled}. The state does not contain the goal -- instead, the agent must explore different locations to locate the goal on its own. The reward at meta-test time is provided when the agent reaches within a small distance of the goal. Coherent exploration on this family of tasks requires driving to random locations in the world, which requires a coordinated pattern of actions that is difficult to achieve purely with action-space noise. We use full state representation for this task, leaving vision based policies to future work.

\begin{figure}[!h]
    \centering
    \includegraphics[width=0.3\linewidth]{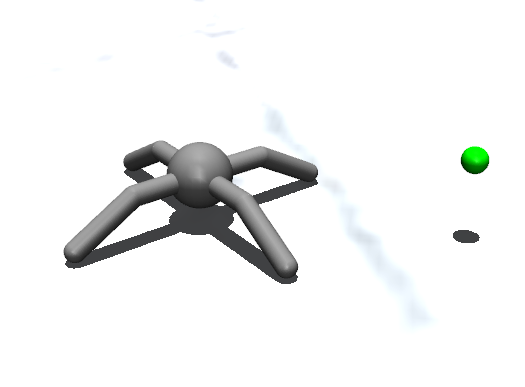}
    \includegraphics[width=0.3\linewidth]{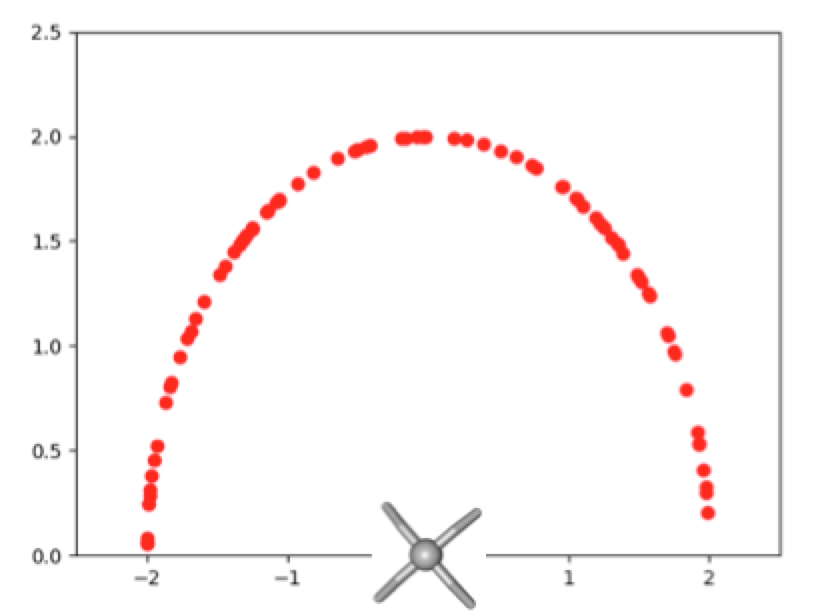}
    \vspace*{-11pt}
    \caption{\textbf{Left:} Legged locomotion with a quadruped (``ant'') navigating to a goal, depicted by the green sphere \textbf{Right:} Depiction of the distribution of tasks. The red points indicate various goals the agent might need to reach, each of which is a task from $p(\tau)$}
\label{fig:ant}
\end{figure}

\paragraph{Legged Locomotion.}
To understand whether we can scale to more complex locomotion tasks with higher dimensionality, the last family of tasks involves a simulated quadruped (``ant'') tasked to walk to randomly placed goals (see Fig.~\ref{fig:ant}), in a similar setup as the wheeled robot. This task presents a further exploration challenge, since only carefully coordinated leg motion actually produces movement to different positions in the world, so an ideal exploration strategy would always walk, but would walk to different places. At meta-training time, the agent receives the negative distance to the goal as the reward, while at meta-test time, reward is only given within a small distance of the goal.

\subsection{Comparisons}
\label{sec:baselines}
We compare our method to a number of prior methods in meta-learning, multi-task learning, learning from scratch, and using task agnostic exploration strategies.
We compare MAESN with RL$^2$ \cite{rl2}, MAML \cite{maml}, and simply learning latent spaces without fast adaptation, analogously to \citet{hausman}.
As discussed in Section~\ref{sec:adaptation}, learning to explore requires actually solving the exploration problem at meta-training, and tasks where exploration is extremely challenging are still challenging at meta-training time. For this reason, all of the methods were provided with dense rewards during meta-training, but meta-testing was still done on sparse rewards. For training from scratch, we compare with TRPO~\cite{TRPO}, REINFORCE~\cite{Williams92}, and training from scratch with VIME~\cite{VIME}. Details of our comparisons and experimental setup can be found in the Appendix.

\subsection{Adaptation to New Tasks}

\begin{figure*}[!t]
\centering
\includegraphics[width=1\linewidth]{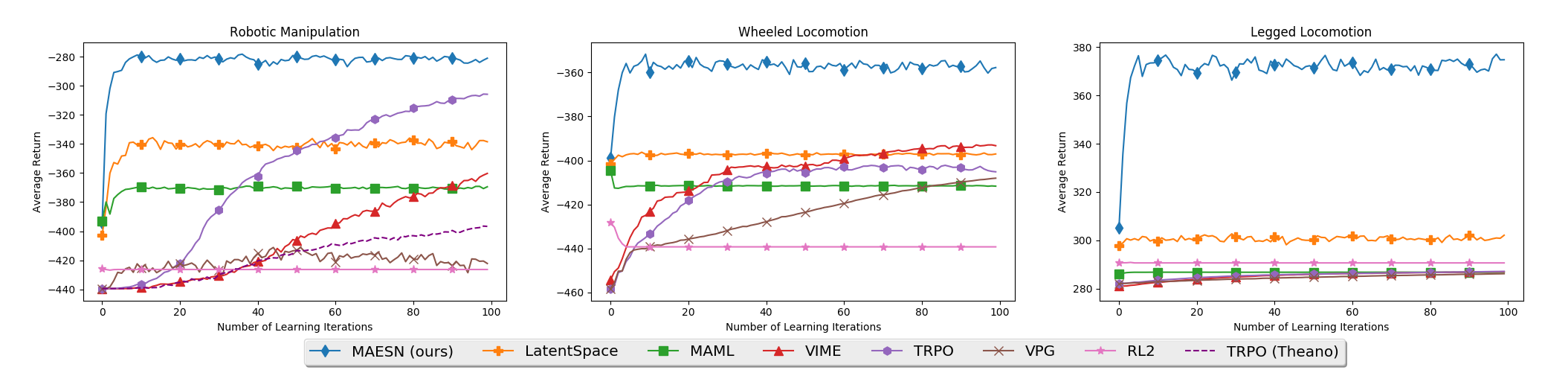}

\caption{Learning progress on novel tasks with sparse rewards for wheeled locomotion, legged locomotion, and object manipulation tasks. The rewards are averaged over 100 validation tasks, which have sparse rewards as described in supplementary materials. We see that MAESN learns significantly better policies, and learns much quicker than prior meta-learning approaches and learning from scratch.On the robotic manipulation task, VIME performs worse than TRPO because the implementation of TRPO it is built on performs poorly. The performance of the base implementation is plotted as TRPO (Theano) for comparison}
\label{fig:plots}
\vspace{-0.2in}
\end{figure*}

We study how quickly we can learn behaviors on new tasks with sparse reward when meta-trained with MAESN, as compared to prior methods. Since the reward function is sparse, success requires good exploration. We plot the performance of all methods in terms of the reward (averaged across 100 validation tasks) that the methods obtain while adapting to tasks drawn from a validation set in Figure~\ref{fig:plots}.
Our results on the three tasks we discussed above show that MAESN is able to explore and adapt quickly on sparse reward environments. In comparison, MAML and RL$^2$ don't learn behaviors that explore as effectively. The pure latent spaces model (LatentSpace in Figure~\ref{fig:plots}) achieves reasonable performance, but is limited in terms of its capacity to improve beyond the initial identification of latent space parameters and is not optimized for fast gradient-based adaptation in the latent space. Since MAESN can train the latent space explicitly for fast adaptation, it can achieve better results faster.

We also observe that, for many tasks, learning from scratch actually provides a competitive baseline to prior meta-learning methods in terms of asymptotic performance. This indicates that the task distributions are quite challenging, and simply memorizing the meta-training tasks is insufficient to succeed. However, in all cases, we see that MAESN is able to outperform learning from scratch and task-agnostic exploration in terms of both learning speed and asymptotic performance. On the manipulation task, learning from scratch is the strongest alternative method, achieving asymptotic performance that is close to MAESN, but more slowly. On the challenging legged locomotion task, which requires coherent walking behaviors to random locations in the world to discover the sparse rewards, we find that only MAESN is able to adapt effectively.

\subsection{Exploration Strategies}

\begin{figure}[!h]
\centering
\begin{tabular}{ccc}
MAESN & MAML & Random \\[6pt]
    \includegraphics[scale = 0.2]{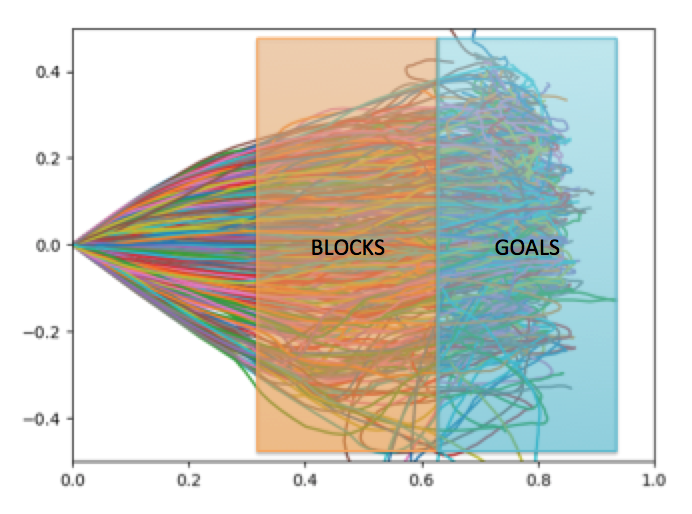} & \includegraphics[scale = 0.17]{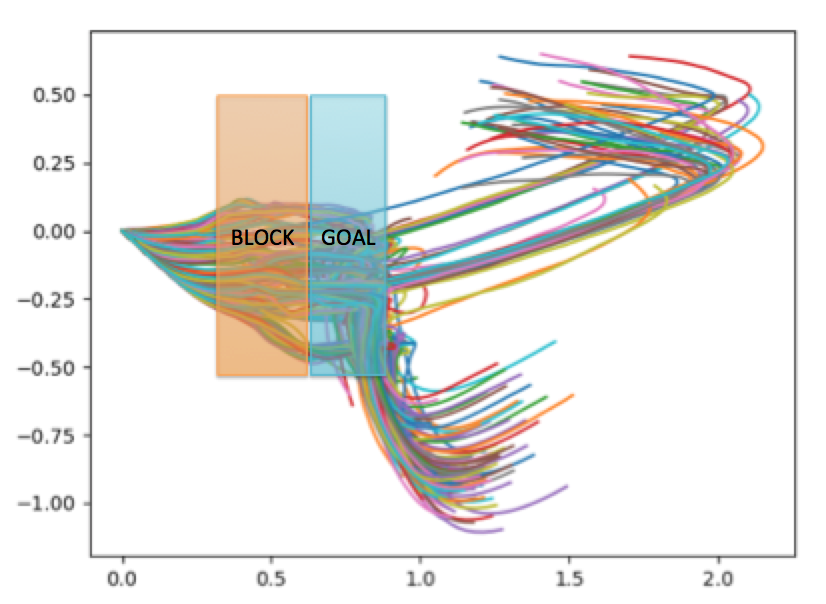} & \includegraphics[scale = 0.2]{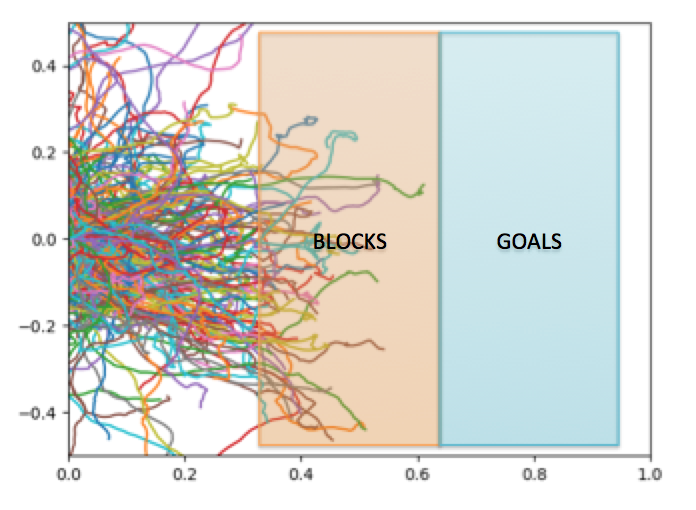}\\
    \includegraphics[scale = 0.2]{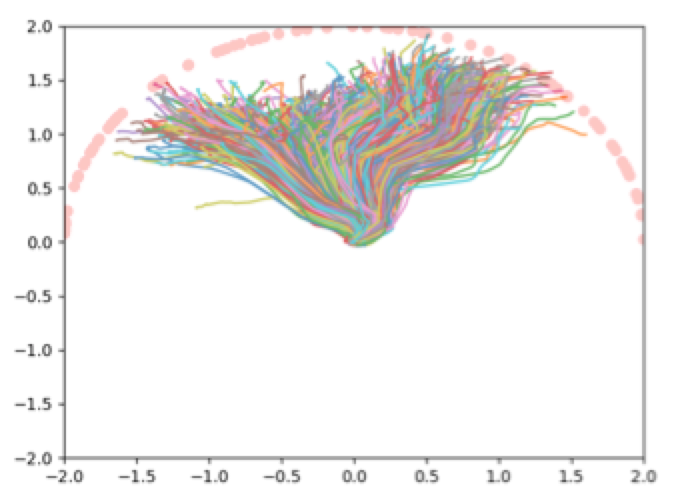} & \includegraphics[scale = 0.2]{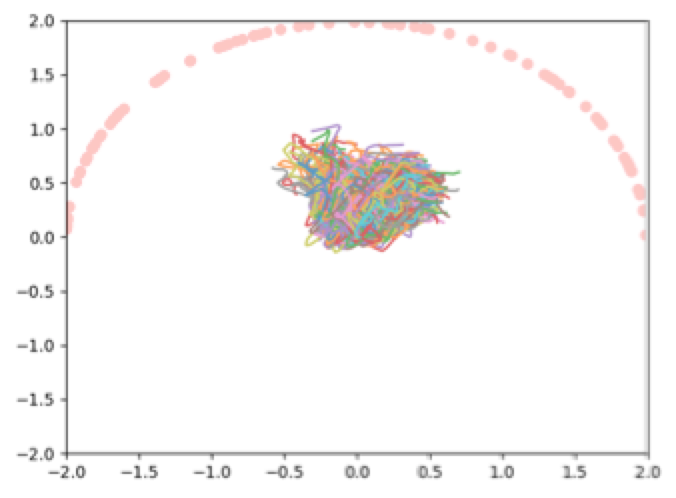} & \includegraphics[scale = 0.2]{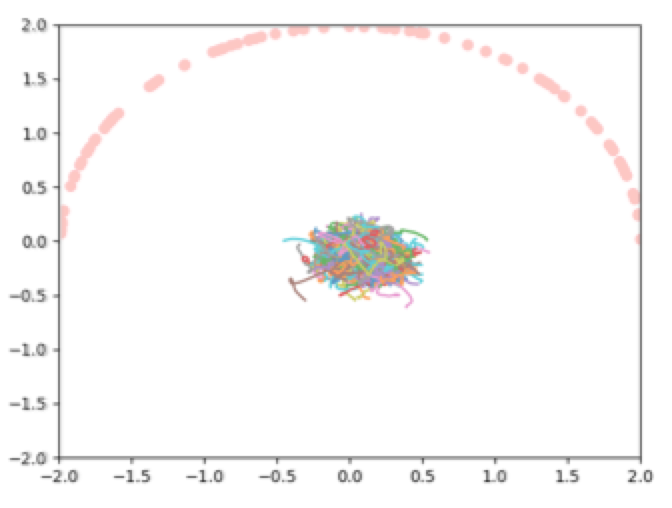}\\
    \includegraphics[scale = 0.2]{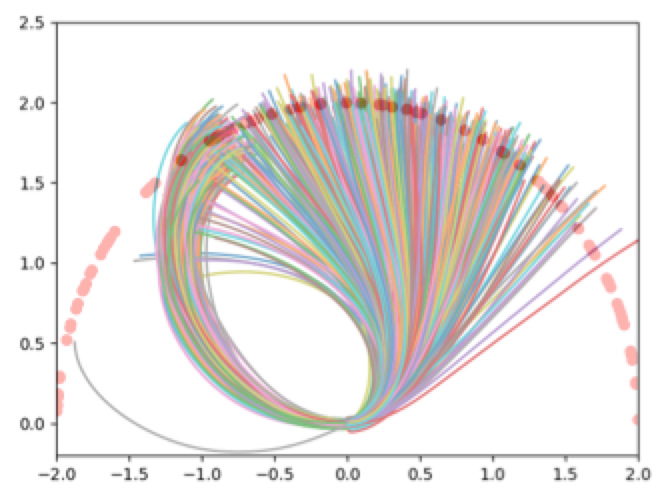} & \includegraphics[scale = 0.2]{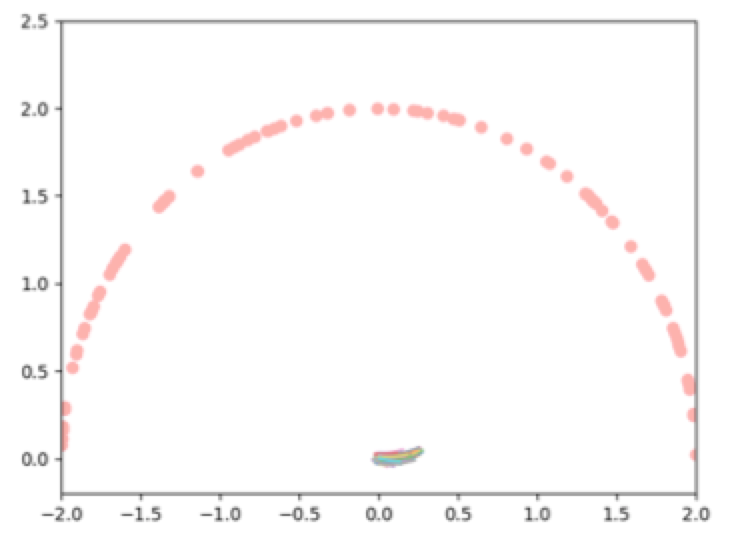} & \includegraphics[scale = 0.2]{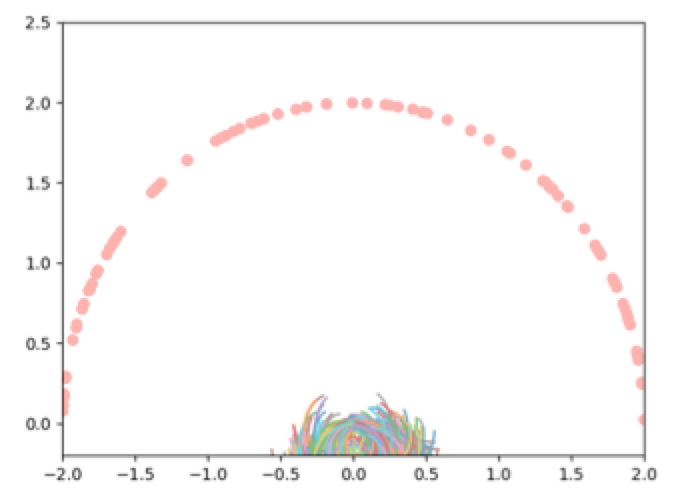}\\
\end{tabular}
\caption{Plot of exploration behavior visualizing 2D position of the manipulator (for blockpushing) and the CoM for locomotion for MAESN, MAML and random initialization. \textbf{Top:} Block Manipulation \textbf{Middle:} Ant Locomotion \textbf{Bottom:} Wheeled Locomotion. Goals are indicated by the translucent overlays. We see that MAESN better captures the task distribution than other methods.}
\vspace*{-10pt}
\label{fig:exploretrajs}
\end{figure}

To better understand the types of exploration strategies learned by MAESN, we visualize the trajectories obtained by sampling from the meta-learned latent-conditioned policy $\pi_{\theta}$ with the latent distribution $\mathcal{N}(\mu, \sigma)$ set to the prior $\mathcal{N}(0, I)$. The resulting trajectories show the 2D position of the hand for the block pushing task and the 2D position of the center of mass for the locomotion tasks.
The task distributions $p(\tau)$ for each family of tasks are shown in Fig~\ref{fig:blockpush},~\ref{fig:wheeled},~\ref{fig:ant}. 
We can see from these trajectories (Fig~\ref{fig:exploretrajs}) that learned exploration strategies explore in the space of coherent behaviors broadly and effectively, especially in comparison with random exploration and standard MAML.

\subsection{Latent Space Structure}

\begin{figure}
\centering
\includegraphics[scale = 0.15]{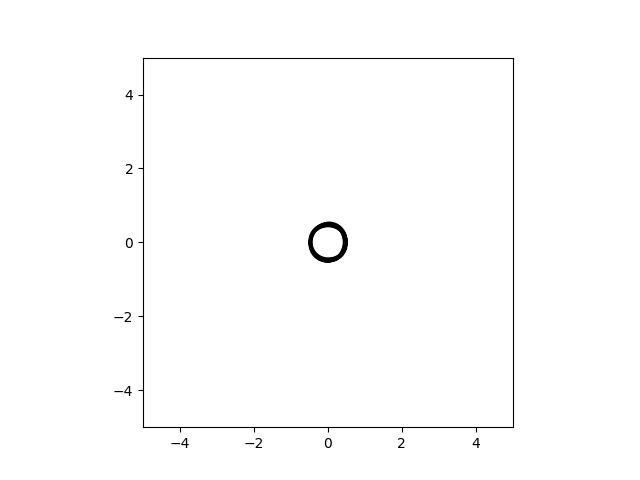}
\includegraphics[scale = 0.15]{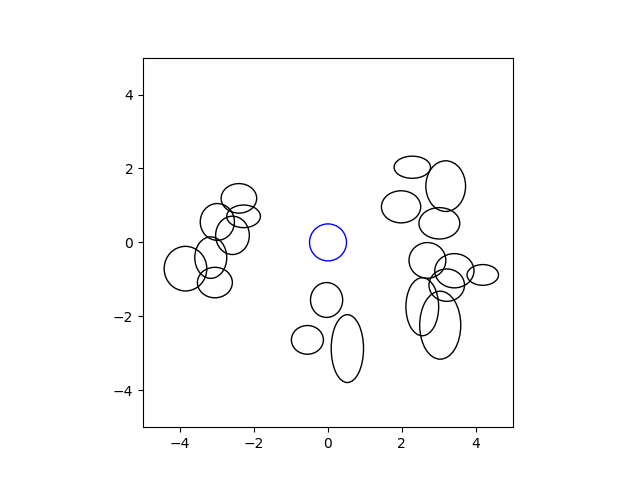}
\caption{Block manipulation latent distributions update in MAESN visualized for a 2D latent space. The ellipses represent the latent distributions $\mathcal{N}(\mu_i, \sigma_i)$ \textbf{Left:} Pre-update latents, \textbf{Right:} Post update latents.  Pre-update latents are driven to the prior, adapted to different post update latents by policy gradient}
\label{fig:latents}
\vspace*{-10pt}
\end{figure}

We investigate the structure of the learned latent space in the object manipulation task by visualizing pre-update $(\mu_i, \sigma_i)$ and post-update $(\mu_i', \sigma_i')$ parameters for a 2D latent space. The variational distributions are plotted as ellipses in this 2D space. As can be seen from Fig~\ref{fig:latents}, the pre-update parameters are all driven to the prior $\mathcal{N}(0, I)$, while the post-update parameters move to different locations in the latent space to adapt to their respective tasks. This indicates that the meta-training process effectively utilizes the latent variables, but also effectively minimizes the KL-divergence against the prior, ensuring that initializing $(\mu,\sigma)$ to the prior for a new task will still produce effective exploration.

\subsection{Role of Structured Noise}

\begin{figure}
      \centering
   \includegraphics[width=0.3\linewidth]{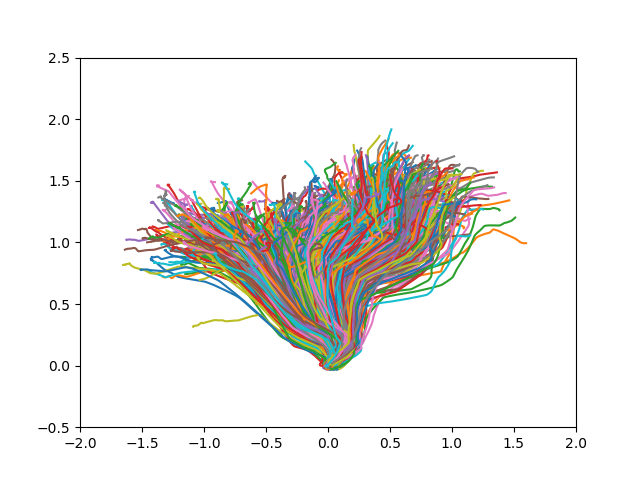}
   \includegraphics[width=0.3\linewidth]{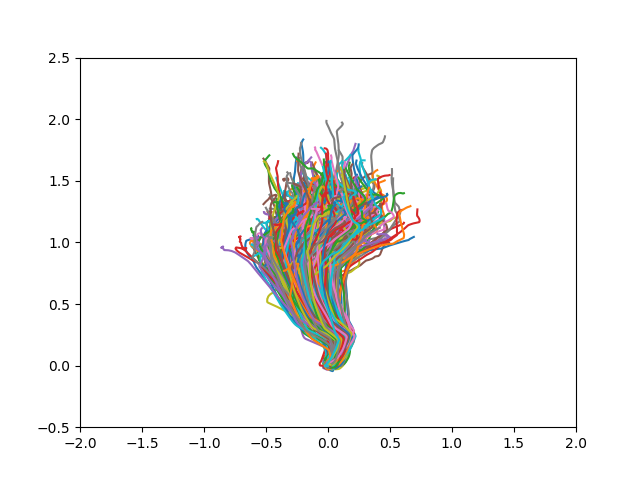}
    \caption{Role of structured noise in exploration with MAESN for the ant legged robot. \textbf{Left:} CoM visitations using structured noise. \textbf{Right:} CoM visitations with no structured noise. Increased spread of exploration and wider trajectory distribution suggests that structured noise \emph{is} being used.}
    \label{fig:zeronoise}
    \vspace*{-10pt}
\end{figure}

To better understand the importance of components of MAESN, we evaluate whether the noise injected from the latent space learned by MAESN is actually used for exploration. We observe the exploratory behavior displayed by a policy trained with MAESN when the latent variable $z$ is kept fixed, as compared to when it is sampled from the learned latent distribution. We can see from Fig.~\ref{fig:zeronoise} that, although there is some random exploration even without latent space sampling, the range of trajectories is substantially broader when $z$ is sampled from the prior. A more detailed ablation study can be found in the supplementary materials.

\section{Conclusion}

We presented MAESN, a meta-RL algorithm that explicitly learns to explore by combining gradient-based meta-learning with a learned latent exploration space. MAESN learns a latent space that can be used to inject temporally correlated, coherent stochasticity into the policy to explore effectively at meta-test time. An intelligent and coherent exploration strategy must randomly sample from among the \emph{useful} behaviors, while omitting behaviors that are never useful. Our experimental evaluation illustrates that MAESN does precisely this, outperforming both prior meta-learning methods and RL algorithms that learn from scratch, including methods that use task-agnostic exploration strategies based on intrinsic motivation~\cite{VIME}. It's worth noting, however, that our approach is not mutually exclusive with intrinsic motivation, and in fact a promising direction for future work would be to combine our approach with novelty based exploration methods such as VIME~\cite{VIME} and pseudocount-based exploration~\cite{pseudocounts}, potentially with meta-learning aiding in the acquisition of effective intrinsic motivation strategies.

\section{Acknowledgements}
The authors would like to thank Chelsea Finn, Gregory Kahn, Ignasi Clavera for thoughtful discussions and Justin Fu, Marvin Zhang for comments on an early version of the paper. This work was supported by a National Science Foundation Graduate Research Fellowship for Abhishek Gupta, ONR PECASE award for Pieter Abbeel, and the National Science Foundation through IIS-1651843 and IIS-1614653, as well as an ONR Young Investigator Program award for Sergey Levine. 

\bibliography{references}
\bibliographystyle{icml2018}

\appendix
\section{Experimental Details}
We built all of our implementation on the open source implementation of rllab~\cite{rllab} and MAML~\cite{maml}. Our policies were all feedforward policies of 2 layers, with a hundred units each and ReLU nonlinearities. We performed meta-training for a single step of adaptation, though longer could be done in principle. 

We found that to get MAESN to work well, meta-learning a per-parameter stepsize is crucial, rather than keeping step-size fixed. This has been found to help in prior work~\cite{metastep} as well.

\section{Reward Functions}
While training all tasks we used dense reward functions to enable meta-training as described in Section~\ref{sec:experiments} of the paper. For each of the tasks, the dense rewards are given by
\begin{equation}
R_{block} = -\|x_{obj} - x_{goal}\|_2
\end{equation}
\begin{equation}
R_{wheeled} = -\|x_{com} - x_{goal}\|_2
\end{equation}
\begin{equation}
R_{ant} = -\|x_{com} - x_{goal}\|_2
\end{equation}
The test-time reward is sparser, provided only in a region around the target position. The sparse rewards for these tasks are given by 
\[ R_{block} = \begin{cases} 
      -c_{max} & \|x_{obj} - x_{goal}\|_2 > 0.2 \\
      -\|x_{obj} - x_{goal}\|_2 &  \|x_{obj} - x_{goal}\|_2 \leq 0.2 \\
   \end{cases}
\]
\[ R_{wheeled} = \begin{cases} 
      -c_{max} & \|x_{com} - x_{goal}\|_2 > 0.8 \\
      -\|x_{obj} - x_{goal}\|_2 &  \|x_{obj} - x_{goal}\|_2 \leq 0.8 \\
   \end{cases}
\]
\[ R_{ant} = \begin{cases} 
      4 -c_{max} & \|x_{obj} - x_{goal}\|_2 > 0.8 \\
      4 -\|x_{obj} - x_{goal}\|_2 &  \|x_{obj} - x_{goal}\|_2 \leq 0.8 \\
   \end{cases}
\]
where $-c_{max}$ is an uninformative large negative constant reward. The reward is uninformative until the agent/object reach a threshold distance around the goal, and then the negative distance to the goal is subsequently provided as the reward function. 

\section{Ablation Study}
Since MAESN introduces a number of components such as adaptive step size, a latent space for exploration to the framework of MAML, we perform ablations to see which of these make a major difference. Adding in a learned latent space (called bias transformation) has been explored before in ~\cite{oneshotimitationchelsea} but the latent space was not stochastic, making it non-helpful for exploration. 

\begin{figure}[!h]
    \centering
    \includegraphics[width=0.8\linewidth]{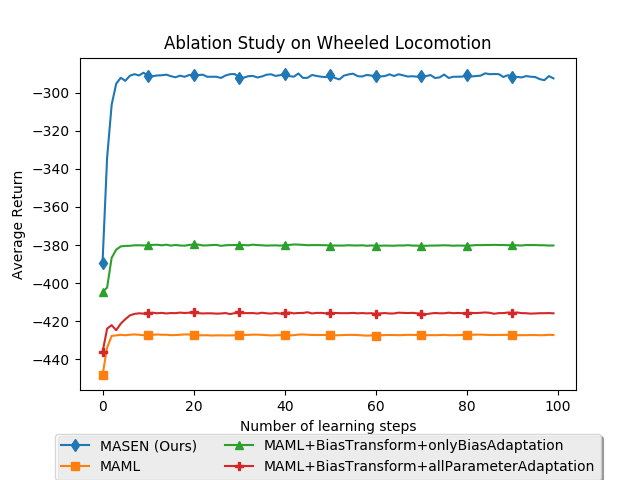}
    \caption{Ablation study comparing adaptation performance on novel tasks of MAESN against a number of variants of MAML - using bias transformation, adaptive stepsize or a combination of both}
    \label{fig:ablations}
\end{figure}

We found that although adding in a bias transformation to MAML was helpful, it did not match the performance of MAESN. Variants considered are (1) standard MAML (2) MAML + bias transform + adaptive stepsize, adapting all parameters in the inner update (maml+Bias +allParameterAdaptation) (3) MAML + bias transform + adaptive stepsize, adapting only the bias parameters in the inner update (maml+Bias +onlyBiasAdaptation). 

\end{document}